\documentclass[letterpaper, 10 pt, conference]{ieeeconf}  

\IEEEoverridecommandlockouts                              

\usepackage{graphics} 
\usepackage{times} 
\usepackage{amsmath} 
\usepackage{amssymb}  
\usepackage{bbold}
\usepackage{arydshln}
\usepackage{multirow}
\usepackage{booktabs} 

\usepackage{xcolor}
\usepackage{soul}

\usepackage[nobiblatex]{xurl}

\usepackage{physics}

\usepackage{pifont}

\usepackage{adjustbox}

\usepackage{hyperref}

\usepackage{graphicx}

\let\labelindent\IEEElabelindent
\usepackage{enumitem}

\definecolor{fei}{rgb}{0.0, 0.0, 0.0}

\title{\LARGE \bf
SuPerPM: A Surgical Perception Framework Based on Deep Point Matching Learned from Physical Constrained Simulation Data
}

\author{Shan Lin$^1$, Albert J. Miao$^1$, Ali Alabiad$^1$, Fei Liu$^1$, Kaiyuan Wang$^1$, \\Jingpei Lu$^1$, Florian Richter$^1$, Michael C. Yip$^1$, \textit{Senior Member, IEEE} 
\thanks{{}$^1$S. Lin, A.J. Miao, A. Alabiad, F. Liu, K. Wang, J. Lu, F. Richter, and M.C. Yip are with the Department of Electrical and Computer Engineering, University of California San Diego, La Jolla, CA 92093, USA.
        (e-mail: \{shl102, amiao, aalabiad, f4liu, k5wang, jil360, frichter, yip\}@ucsd.edu) %
}}

\begin{document}
\bstctlcite{IEEEexample:BSTcontrol}

\maketitle
\thispagestyle{empty}
\pagestyle{empty}

\begin{abstract}
A major source of endoscopic tissue tracking errors during deformations stems from wrong data association between observed sensor measurements with previously tracked scene.
To mitigate this issue, we present a surgical perception framework, SuPerPM, that leverages learning-based non-rigid point cloud matching for data association, thus accommodating larger deformations than previous approaches which relied on Iterative Closest Point (ICP) for point associations.
The learning models typically require training data with ground truth point cloud correspondences, which is challenging or even impractical to collect in surgical environments. 
Thus, for tuning the learning model, we gather endoscopic data of soft tissue being manipulated by a surgical robot and then establish correspondences between point clouds at different time points to serve as ground truth. 
This was achieved by employing a position-based dynamics (PBD) simulation to ensure that the correspondences adhered to physical constraints. 
The proposed framework is demonstrated on several challenging surgical datasets that are characterized by large deformations, achieving superior performance over advanced surgical scene tracking algorithms. \footnote{Our data and code are available at \url{https://github.com/ucsdarclab/SuPerPM.git}.}
\end{abstract}


\section{Introduction} \label{sec:intro}
With the growing popularity of endoscopic procedures, more assistive technologies can be integrated into  operating rooms.
Overlaying virtual visualizations from pre-operative scans of anatomy helps surgeons identify sensitive organs during procedures \cite{ukimura2008imaging, schulze2010intra}.
Further help can be done intra-operatively by identifying tissue types directly from the endoscopic image data \cite{prasath2016polyp, grammatikopoulou2021cadis}.
In the case of robotic surgery, automation efforts are being actively researched \cite{yip2019robot, haidegger2019autonomy}.
A foundational technology for these efforts is tissue tracking and reconstruction from endoscopic images.
However, this remains an unsolved challenge, with performance compromised under demanding conditions like poor lighting of the tissue, frequent occlusions, and large tissue deformations. 

\begin{figure}[t]
\centering
\includegraphics[width=0.99\linewidth]{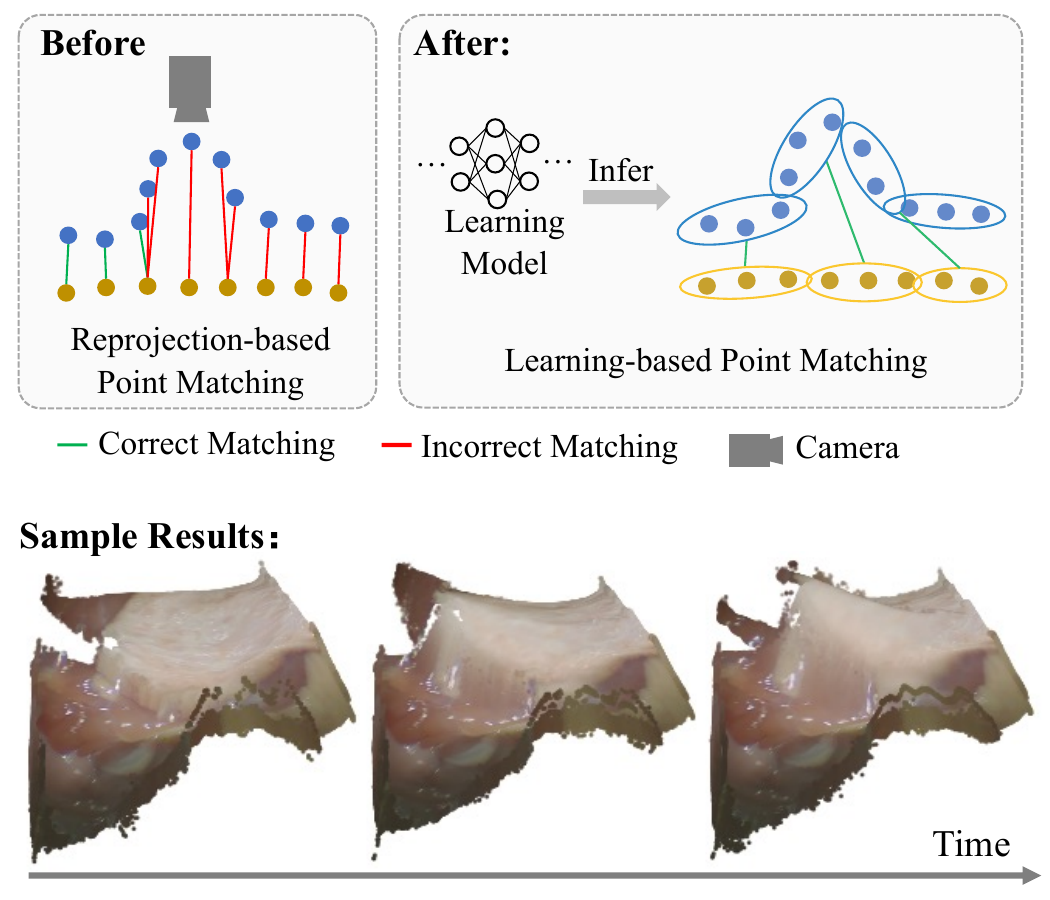}
\vspace{-2em}
\caption{We propose SuPerPM, a perception framework for endoscopic surgical scenes based on deep point matching. The perception framework, SuPer, that we built upon implements tracking based on the ICP algorithm. It projects the points onto the image plane to find the point cloud correspondences and suffers from incorrect associations (Before). We substitute the projection-based matching with a deep learning model (After). It extracts deep features, each covers the information of a certain field of view on the point cloud, and can achieve more accurate association after training. An example of tracking and reconstruction results for a deforming surgical scene is shown in the bottom of the figure. }
\label{fig:cover_fig}
\vspace{-1.5em}
\end{figure}

For endoscopic tissue tracking, the primary challenge arises from the need to establish data associations on soft tissue surfaces, which commonly lack robust features. 
Some methods \cite{li2020super, lu2021super} use the Iterative Closest Point (ICP) algorithm \cite{rusinkiewicz2001efficient, low2004linear} to iteratively match nearest point pairs for estimating transformations. 
However, this greedy search only finds correspondences within local areas, while solving for the true correspondences would require full geometric consideration as shown in Figure \ref{fig:cover_fig}. 
For rigid scenes, all points share the transformation parameters, so the estimation could converge through iteration even with such inaccurate correspondences. 
In contrast, in deforming scenes, each point undergoes distinct transformations, which can cause the ICP algorithm to prematurely ``converge" to incorrect transformations. 
Alternatively, other studies have proposed to use all information in the image for data association (direct methods) \cite{engel2014lsd}, adopting techniques like optical flow and photometric cost \cite{gomez2021sd, lin2022semantic}, or leveraging semantic information to offer additional guidance \cite{lin2022semantic}. 
Yet, these approaches have not resolved the gradient-locality issue, which can still trap the estimation within a local minimum. 

Recently, deep learning models have demonstrated their capability to derive enhanced feature representations from 3D data \cite{deng2021vector, simeonov2022neural, thach2022learning}, thereby enabling more robust point cloud matching \cite{li2022lepard, fu2021robust}. 
This paper is built upon a point-plane ICP cost-based surgical perception framework named \textbf{Su}rgical \textbf{Per}ception (SuPer) \cite{li2020super, lu2021super}. 
We enhance the data association for the ICP cost within our framework, SuPerPM, by harnessing these advancements on learning models. 
Moreover, most learning-based point cloud matching models require at least sparsely annotated correspondences between point clouds for training. 
However, procuring dense and accurate ground truth correspondences for surgical scenes is quite challenging. 
Thus, we propose a novel pipeline for generating deformed point cloud pairs for fine-tuning the point cloud matching model. This pipeline leverages the position-based dynamics (PBD) simulation \cite{Bender_2017, Fei_2021_ICRA}, which formulates physical constraints with positional and geometric data. PBD is capable of real-time simulations, enabling the dynamic deformation of objects. Consequently, we can establish a direct connection for mapping the physical simulation to point-cloud perception, facilitating point-wise positional mapping.

\begin{figure*}[!t]
\centering
\includegraphics[width=0.99\linewidth]{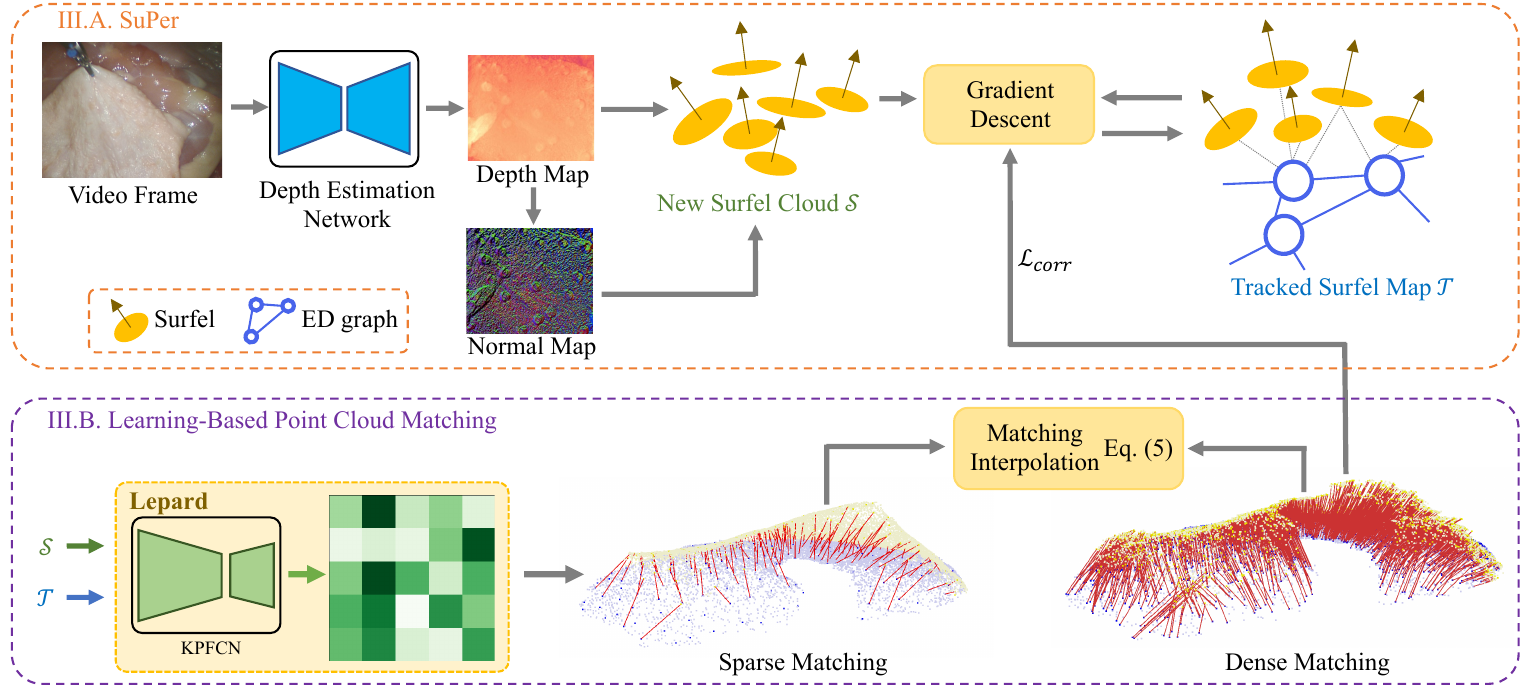}
\vspace{-0.5em}
\caption{Overview of SuPerPM. A learning-based point cloud matching method called Lepard \cite{li2022lepard} is integrated (Section \ref{sec:lepard}) into the surgical perception framework SuPer (Section \ref{sec:super}). 
By leveraging a data-driven method for making point-wise associations, SuPerPM is robust against poor point association that conventional approaches using ICP run into.
Lepard is finetuned using synthetically generated point cloud pairs obtained using PBD simulation as described in Section \ref{sec:syn}.} \label{fig:overview}
\vspace{-1em}
\end{figure*}

In summary, the main contributions are as follows: 
\begin{itemize}
\item Integrate a learning-based non-rigid point cloud matching method into a surgical perception framework to improve data association for tissue tracking. 
\item \textcolor{fei}{Propose a pipeline for synthesizing non-rigid point cloud pairs using a physical constraints-based simulator (i.e., PBD), to facilitate fine-tuning of the learning-based point cloud matching method. }
\item Release a robotic tissue manipulation dataset with large deformations collected using the da Vinci Research Kit (dVRK) \cite{kazanzides2014open}.
\item Conduct extensive experiments on public and newly collected endoscopic data and demonstrate the performance of the proposed framework. 
\end{itemize}

\section{Related Work}


\subsection{Endoscopic Tissue Tracking and Reconstruction} \label{sec:lit_tissue_tracking}
Endoscopic tissue tracking is a specialized domain within non-rigid tracking, presenting significant challenges due to the deformable nature of the tissue. 
A branch of approaches relies on the ICP algorithm, which iteratively identifies nearest point pairs in the Euclidean or geodesic space for transformation estimation \cite{gomez2021sd, recasens2021endo, li2020super, lu2021super}. 
However, for the deformable object, each point has a distinct transformation, and the relationships between the transformations of adjacent points are much weaker than in a rigid object. Therefore, obtaining accurate matches in early optimization iterations becomes crucial; otherwise, the transformations can quickly adapt to the wrong matches. 
To enhance data association, some works adopt direct methods \cite{engel2014lsd} that utilize dense image information, such as photometric loss \cite{gomez2021sd, recasens2021endo}. Others resort to integrating additional data modalities like semantic information to guide data association \cite{lin2022semantic}. 
Moreover, together with the aforementioned strategies, existing works typically employ regularization terms based on rigidness assumptions to mitigate the impact of noisy associations. 
This includes the use of the as-rigid-as-possible (ASAP) cost to ensure neighboring points move in close proximity to each other \cite{grasa2011ekf, grasa2013visual, marmol2019dense}. 
Yet, these regularization terms can only partially address performance degradation from incorrect data associations. 
In this work, we focus on further improving data association by learning-based point cloud matching. 

\subsection{Non-rigid Point Cloud Matching} 
Non-rigid matching between deformed point clouds is the key that influences our tissue tracking and reconstruction performance. 
In addition to surgical applications, precise point cloud matching is also crucial for many other tasks involving non-rigid objects. 
On top of ICP or photometric costs, many methods also incorporate visual features like SIFT \cite{lowe2004distinctive} to provide additional correspondence information \cite{innmann2016volumedeform}. 
However, these conventional features are known to lack robustness for surgical scenes involving texture-less and moistened tissues. 
Recently, learning-based models have demonstrated their superior performance in representation learning for 3D data \cite{deng2021vector, simeonov2022neural} and identifying correspondences between data \cite{li2022lepard, groueix20183d, niemeyer2019occupancy, bozic2020deepdeform}, showing their advantages for non-rigid registration \cite{li2020learning, bozic2020neural}. 
We leverage these recent advances in learning-based non-rigid matching to provide better correspondences for tissue tracking under large deformations. 

\section{Method}
The proposed method is developed on our previous work SuPer \cite{li2020super, lu2021super}. 
In Section \ref{sec:super}, we provide a brief recap of SuPer. Then, in Section \ref{sec:lepard}, we describe how we integrate a learning-based point matching model, Lepard \cite{li2022lepard}, into SuPer to enhance its data association. 
\textcolor{fei}{Surgical scenes often have flat tissue surfaces with fewer distinct features, unlike objects (e.g., animals in DeformingThings4D \cite{li20214dcomplete}) that can be found in many public datasets for non-rigid registration. Therefore, for fine-tuning Lepard, we present a pipeline designed to establish ground truth correspondences between deformed point cloud pairs by employing the position-based dynamics (PBD) simulation framework, ensuring the adherence to physical constraints, as outlined in Section \ref{sec:syn}. }
The overview of the proposed framework is shown in Figure \ref{fig:overview}.

\subsection{SuPer Framework} \label{sec:super}
SuPer performs reconstruction and tracking of the entire scene, encompassing both the surgical tools and the deforming soft tissues. This study focuses on tissue tracking, for further insights and details on other aspects of SuPer, please see \cite{li2020super, 9565398}. 

\subsubsection{Scene Representation}
In SuPer, the tissue is tracked with a model-free method and is represented using surface elements (surfels) \cite{keller2013real, gao2019surfelwarp}. Each surfel $\mathcal{S}$ is defined by a position $\vb{p}_i\in\mathbb{R}^3$, a normal $\vb{n}_i\in\mathbb{R}^3$, a color $\vb{c}_i\in\mathbb{R}^3$, a radius $\mathbb{r}_i\in\mathbb{R}$, a confidence score $\mathbb{c}_i\in\mathbb{R}$, and a timestamp $\mathbb{t}_i\in\mathbb{N}$ of when it was last updated. 
The quantity of surfels is directly linked to the number of image pixels, which can result in substantial computational requirements when estimating individual surfel transformations. 
To address this issue, SuPer employs the Embedded Deformation (ED) graph \cite{sumner2007embedded} that has sparser vertices (named ED nodes) to drive surfels' motions. 
The ED graph consists of a set of vertices $\mathcal{V}$, a set of edges $\mathcal{E}$, and a set of parameters $\Gamma$, \emph{i.e.}, $\mathcal{G}_{ED}=\{\mathcal{V}, \mathcal{E}, \Gamma\}$. The parameters for each ED node are defined as $(\vb{g}_j, \vb{q}_j, \vb{b}_j)\in\Gamma$, where $\vb{g}_j\in\mathbb{R}^3$ is the position, $\vb{q}_j\in\mathbb{R}^4$ and $\vb{b}_j\in\mathbb{R}^3$ are the quaternion and translation parameters.

\subsubsection{Transformation Estimation}
At every new video frame, a total of $7\times(|\mathcal{V}|+1)$ parameters ($|\mathcal{V}|$ is the number of ED nodes) will be estimated, \emph{i.e.}, $\vb{q}_j$ and $\vb{b}_j$ for each ED node, and a global homogeneous transformation matrix $\vb{T}_g \in SE(3)$ shared by all surfels. 
The parameters are optimized by minimizing a total cost function (see Section \ref{sec:cost}) using gradient descent \cite{sratradeoffs} with PyTorch’s automatic differentiation. 
Subsequently, the estimated parameters are utilized to update the surfel positions and normals: 
\begin{equation} \label{update_p}
\widetilde{\overline{\vb{p}}_i} = \vb{T}_g\sum_{j\in \mathcal{N}(\vb{p}_i)}\omega_j(\vb{p}_i)[T(\vb{q}_j, \vb{b}_j)(\overline{\vb{p}}_i-\vec{\vb{g}}_j)+\vec{\vb{g}}_j]
\end{equation}
\begin{equation} \label{update_n}
\widetilde{\vec{\vb{n}}_i} = \vb{T}_g\sum_{j\in \mathcal{N}(\vb{p}_i)}\omega_j(\vb{p}_i)[T(\vb{q}_j, 0)\vec{\vb{n}}_i]
\end{equation}
where $T(\vb{q}_j, \vb{b}_j) \in SE(3)$ is the homogeneous transform matrix of the $j$th ED node, $\overline{\cdot}$ and $\vec{\cdot}$ are the homogeneous representations of a point and motion, \emph{i.e.} $\overline{\vb{p}}=[\vb{p}, 1]^T$ and $\vec{\vb{g}}=[\vb{g}, 0]^T$. 
$\mathcal{N}(\vb{p}_i)$ is the set of $k$-nearest neighbors (KNN) of $\vb{p}_i$ in $\mathcal{G}_{ED}$ and is re-determined each time new positions and normals of the ED nodes and surfels are obtained. 
$\omega_j({\vb{p}_i})$ is a normalized weight that indicates the influence of $\vb{g}_j$ to $\vb{p}_i$ and is defined as $\omega_j(\vb{p}_i)=\frac{e^{-\|\vb{p}_i-\vb{g}_j\|}}{\sum_{j\in \mathcal{N}_i}e^{-\|\vb{p}_i-\vb{g}_j\|}}$. 
Equations (\ref{update_p}) and (\ref{update_n}) can be interpreted as that the surfels are transformed with the average motion of ED nodes near them. 

\subsubsection{Cost Functions} \label{sec:cost}
The total cost function is given by
\begin{equation}
\label{eq:optimization}
\mathop{\arg \min}\limits_{\vb{q}, \vb{b}, \vb{T}_g} \lambda_{icp}\mathcal{L}_{icp} + \lambda_r\mathcal{L}_{reg}
\end{equation}
where 
$\mathcal{L}_{icp}$ is the point-to-plane ICP cost \cite{low2004linear} that measures the similarity between the tracked data and the new observations, $\mathcal{L}_{reg}$ is the regularization term, $\lambda_{icp}$ and $\lambda_{r}$ are the hyper-parameters. 

The point-to-plane ICP cost is calculated by:
\begin{equation}
\mathcal{L}_{icp} = \sum_{i} (\vec{\vb{n}}_o^T(\widetilde{\overline{\vb{p}}_i}-\overline{\vb{p}}_o))^2
\end{equation}
where $\widetilde{\overline{\vb{p}}_i}$ is a surfel from the tracked surfel cloud, and it is transformed using the currently estimated transformations of the ED nodes. 
To establish the correspondence between $\widetilde{\overline{\vb{p}}_i}$ and the new data, we project $\widetilde{\overline{\vb{p}}_i}$ onto the new image plane and conduct bilinear sampling \cite{jaderberg2015spatial} on the depth and normal maps to acquire the corresponding position and normal observations $\overline{\vb{p}}_o$ and $\vec{\vb{n}}_o$. 
As illustrated in Section \ref{sec:intro} and \ref{sec:lit_tissue_tracking}, during the initial iterations, these projective correspondences furnish inaccurate information that can lead the transformations toward a local minimum. 
In this work, we propose to mitigate this issue by replacing the ICP cost with an advanced learning-based approach to ensure more accurate associations can established from the beginning. 

The regularization term is composed of two costs. 
One is the as-rigid-as-possible cost that enforces similar movement among neighboring ED nodes. This cost can partially mitigate the effects of incorrect data associations on the ICP algorithm. 
The second cost aims to ensure the estimated quaternions hold $\|\vb{q}\|^2=1$. 
More details and the involved equations can be found in \cite{li2020super}. 

\subsection{Learning-based Point Cloud Matching} \label{sec:lepard}

To enhance data association, we adapt an advanced learning-based point cloud matching method named Lepard \cite{li2022lepard}. 
Lepard first extracts multi-level geometric features using a convolutional backbone designed for point clouds. It then uses a transformer block to enhance these features. The transformer block consists of two layers: a self-attention layer, which aggregates global context, and a cross-attention layer, which interchanges information between the source and target point clouds. 
Next, the aggregated features from the two input point clouds are compared to generate a confidence matrix $\mathcal{C}$, where each element $\mathcal{C}_{i, j}$ represents the confidence level that the corresponding points in the two point clouds are a match. Finally, matches with higher confidence values are selected as the output matches. 

Specifically, Lepard takes surfel positions $\vb{U}\in\mathbb{R}^{N\times3}$ of a source surfel cloud (\emph{i.e.}, the tracked surfel cloud) and the surfel positions $\vb{V}\in\mathbb{R}^{M\times3}$ of a target surfel cloud (\emph{i.e.}, the surfel cloud extracted from the new observations) as input. 
$\vb{U}$ and $\vb{V}$ are first downsampled to $\vb{U}'\in\mathbb{R}^{N'\times3}$ ($N'\ll N$) and $\vb{V}'\in\mathbb{R}^{M'\times3}$ ($M'\ll M$) after passing through a convolutional backbone for point cloud feature extraction, 
resulting in a relatively sparse match set $\mathcal{K}=\{(\vb{u}_1, \vb{v}_1), (\vb{u}_2, \vb{v}_2), ..., (\vb{u}_K, \vb{v}_K)\}$, where $\vb{u}_k\in\mathbb{R}^3$ is a point in $\vb{U}'$ and $\vb{v}_k\in\mathbb{R}^3$ is a point in $\vb{V}'$. 
This downsampling is crucial for extracting complex, hierarchical features from the point cloud for robust matching. It is also necessary for efficient computation, as the subsequent transformer and matching blocks have a computational complexity of $O(n^2)$ \cite{li2022lepard}. 
However, since SuPerPM requires a dense matching between the tracked and new surfel clouds, we conduct interpolation to extend the sparse matches to dense matches. Specifically, for each surfel in the source surfel cloud, we estimate its new position $\hat{\vb{p}}_i$ by averaging the correspondences within the local region of its current position $\vb{p}_i$:
\begin{equation}
\hat{\vb{p}}_i=\vb{p}_i + \sum_{\vb{u}_j\in\mathcal{N}(\vb{p}_i)}\frac{(\vb{v}_j-\vb{u}_j)\|\vb{u}_j-\vb{p}_i\|^{-1}}{\sum_{\vb{u}_k\in\mathcal{N}(\vb{p}_i)}\|\vb{u}_k-\vb{p}_i\|^{-1}}
\end{equation}
where $\mathcal{N}(\vb{p}_i)$ is the set of KNN of $\vb{p}_i$ in $\vb{U}'$. 
Then, the following point-point correspondence cost $\mathcal{L}_{corr}$ is utilized, with weight $\lambda_c$, to substitute the ICP cost $\mathcal{L}_{icp}$ in the total cost function
\begin{equation}
\mathcal{L}_{corr} = \sum_{i} \|\widetilde{\overline{\vb{p}}_i}-\hat{\vb{p}}_i\|^2
\end{equation}

\subsection{\textcolor{fei}{Deformed Point Cloud Pair Synthesis Pipeline}} \label{sec:syn}
\begin{figure}[t]
\centering
\includegraphics[width=0.95\linewidth]{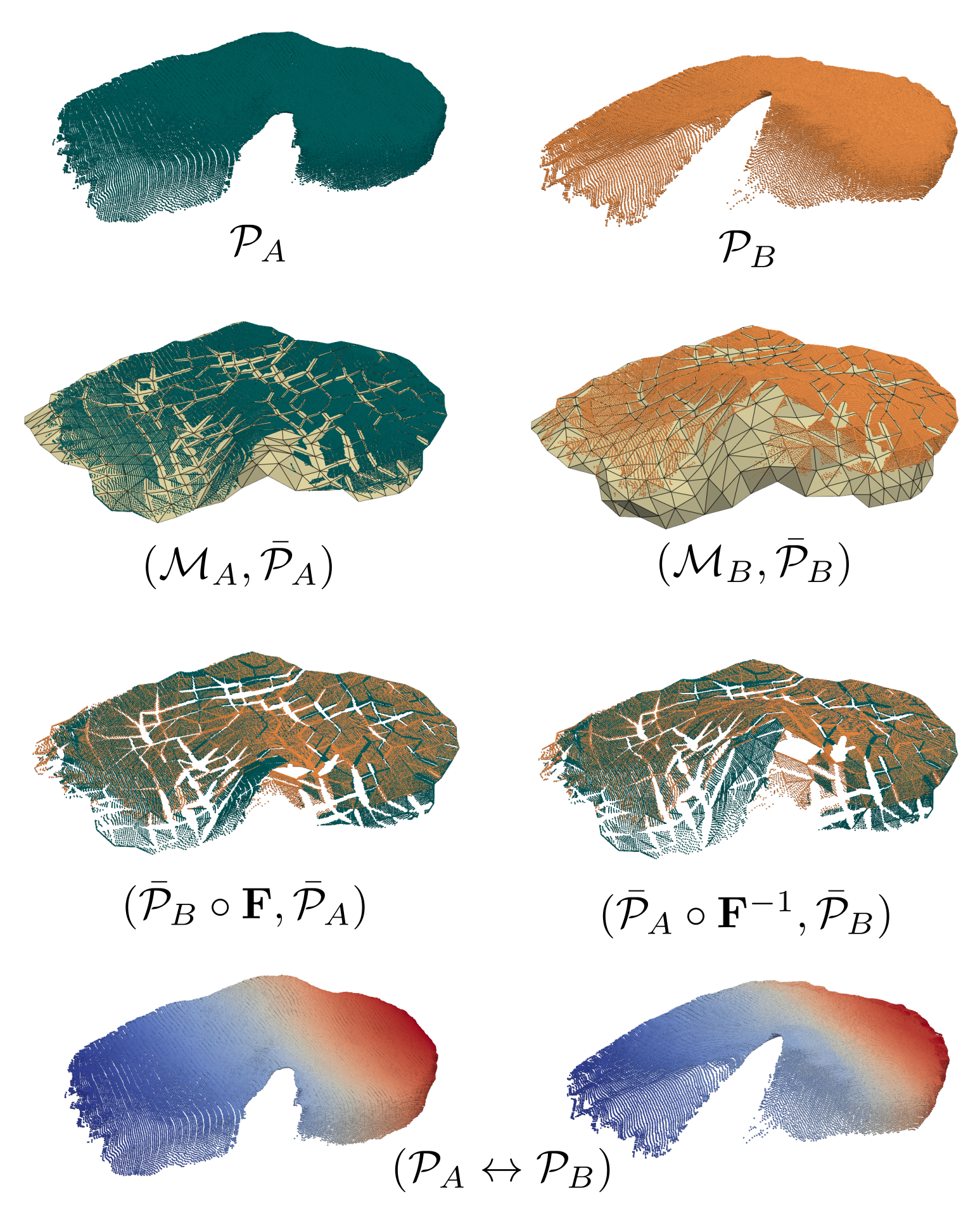}
\vspace{-1em}
\caption{Synthetic Data Generation Pipeline. We generate paired point cloud data to fine-tune Lepard \cite{li2022lepard} on tissue deformations based on real-world data that is registered to a PBD simulation. 
The PBD simulation ensures the associations are physically feasible. }
\label{fig:paired_point_clouds}
\vspace{-1.5em}
\end{figure}

We have adopted the methodology introduced in \cite{Fei_2021_ICRA}, which utilizes position-based dynamics (PBD) for the formulation of physics-based constraints. These constraints, encompassing aspects like volume, distance, and shape matching, are crucial for ensuring stability and revealing the inherent physical properties of the system. This approach is employed to generate a sequence of simulated surface meshes, denoted as $\mathcal{M}$, each containing a set of triangle cells and vertices.
Using the simulated surface meshes that is registered to the real world, our objective is to produce paired point clouds frames $\mathcal{P}_A$ and $\mathcal{P}_B$, reflecting the surface deformations induced by a physics simulation.
To do so, we first project the point cloud data onto the surface mesh by finding the nearest point on $\mathcal{M}_A$ and $\mathcal{M}_B$,
\begin{equation}
\begin{split}
    & \text{Projected } \mathcal{\bar{P}}_A = \arg\min_{\mathbf{v} \in \mathcal{M}_A} \left\| \mathcal{P}_A - \mathbf{v} \right\| \\
    & \text{Projected } \mathcal{\bar{P}}_B = \arg\min_{\mathbf{v} \in \mathcal{M}_B} \left\| \mathcal{P}_B - \mathbf{v} \right\|
\end{split}
\end{equation}
where $\mathbf{v}$ are located inside one of the triangle cell of the mesh.
From the projected point cloud, we apply the deformation gradient tensor, $\mathbf{F}$, which is generated from the PBD simulation, and transform the points as follows
\begin{equation}
\begin{split}
    \mathcal{\bar{P}}^*_B  & = \mathcal{\bar{P}}_A \circ \mathbf{F}^{-1} \\
    \mathcal{\bar{P}}^*_A  & = \mathcal{\bar{P}}_B \circ \mathbf{F} \\
    \mathbf{F} & = \nabla_{_{\mathcal{M}_A}} \mathcal{M}_B
\end{split}
\end{equation}
Hence, we can create the paired dataset for original point clouds $(\mathcal{P}_A \leftrightarrow \mathcal{P}_B)$ by first obtaining the indices of paired transformed points as follows:
\begin{equation}
    \text{Paired points index} = \left\{(\mathcal{\bar{P}}^*_B, \mathcal{\bar{P}}_B) \bigcup (\mathcal{\bar{P}}^*_A, \mathcal{\bar{P}}_A)\right\}
\end{equation}
The whole process pipeline is show in Figure \ref{fig:paired_point_clouds}
It can be seen that the paired points are deformed according to the simulation with physical constraints. 

\section{Experiments and Results}

We demonstrate the proposed framework using the SuPer dataset \cite{li2020super}, which was released alongside the original SuPer framework. Additionally, we introduce a newly collected dataset named SupDef that features larger deformations.

\subsection{Datasets} \label{sec:data}

\subsubsection{SuPer} \label{sec:data_super}
In the SuPer dataset \cite{li2020super}, the da Vinci Research Kit (dVRK) \cite{kazanzides2014open, richter2021bench} was used to control a surgical robotic arm (\emph{i.e.}, Patient Side Manipulator) to mimics the commonly performed tensioning motion in surgery by grasping and tugging a piece of chicken tissue, generating deformations of the scene.  
A single trial (named SuPerV1 in the following sections) that consists of 520 rectified 640$\times$480 video frames was annotated for evaluation.
For evaluation, a total of 20 points on the tissue surfaces were chosen, and their deformation trajectories were annotated throughout the trial. We project the tracked surfels onto the image plane and compare the reprojections with their corresponding ground truth positions to calculate the reprojection errors.

\subsubsection{SupDef}
\begin{figure}[t]
\vspace{0.5em}
\centering
\includegraphics[width=\linewidth]{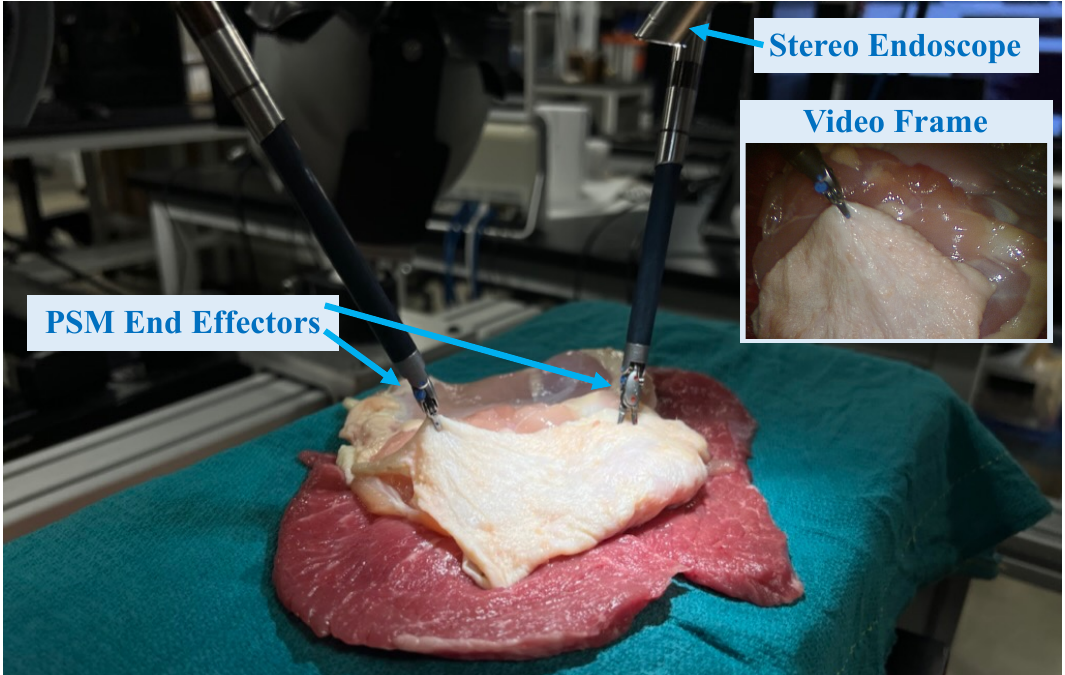}
\vspace{-1.5em}
\caption{Experimental Setup with the dVRK System \cite{kazanzides2014open}. Two PSM arms are commanded to tension and deform the tissue. Meanwhile the stereo endoscope is used to provide the image data as inputs for SuPerPM.}
\label{fig:exp_setup}
\vspace{-1.5em}
\end{figure}

The deformations of tissues in existing public datasets are still fairly small. To demonstrate the benefits of SuPerPM in handling large deformations, we collected a new dataset named SupDef with relatively larger deformations across the entire manipulated tissue. The experiment setup (see Figure \ref{fig:exp_setup}) and method for data postprocessing were identical to the SuPer dataset. 
The primary difference between SupDef and SuPer datasets is that the dVRK was controlled to pull the chicken tissue further away and induce larger deformations. We recorded two manipulation trials, referred to as SuperDef-T1 and SuPerDef-T2, each consisting of about 100 to 200 640$\times$480 rectified video frames captured at 30 fps for tissue tracking. We manually annotated the trajectories of around 10$\sim$20 selected points that undergo large deformation on the tissue surface to serve as ground truth for evaluation.

\subsection{Evaluation Metrics} \label{metrics}

As detailed in Section \ref{sec:data}, all datasets include annotated trajectories of several selected tissue surface points as ground truth. 
We evaluate the tracking algorithms using the average reprojection error to, defined as

\begin{equation}
    e = \frac{1}{TS}\sum_{t=1}^T\sum_{s=1}^S\|\pi(\vb{p}_s^t) - \vb{y}_s^t\|_2
\end{equation}

\noindent where $\pi(\vb{p}_s^t)$ maps the point $\vb{p}_s^t$ from 3D space onto the image plane, $\vb{y}_s^t$ is the corresponding ground truth position of the projection of $\vb{p}_s^t$ in the image plane. For each trial, $T$ is the total number of video frames and $S$ is the total number of annotated points. We average the distance between the surfel projections and their respective ground truth positions across all annotated points and time steps in each trial.


\begin{table}[t!]
\centering
\caption{Reprojection error comparison on SuPer and SupDef.}
\label{tab:compare_sota}
\setlength\tabcolsep{0.9em}
\begin{tabular}{lccc}
\toprule
\multirow{2}{*}{Method} & \multicolumn{3}{c}{Data} \\ \cmidrule(lr){2-4}
  & SuPerV1 & SupDef-T1 & SupDef-T2 \\
\midrule
DefSLAM \cite{lamarca2020defslam} & 17.1(5.5) & 8.1(4.9) & \textbf{28.0(8.6)} \\
SD-DefSLAM \cite{gomez2021sd} & 27.2(18.0) & 9.7(11.5) & 37.9 (22.7) \\ 
SuPer \cite{li2020super} & 9.2(13.1) & 8.6(11.4) & 40.7(26.7) \\ 
\midrule
SuPerPM (Pre-trained) & 11.1(12.3) & 7.2(8.7) & 43.4(27.0) \\
SuPerPM (Fine-tuned) & \textbf{7.9(13.1)} & \textbf{6.2(9.2)} & 34.5(23.6) \\
\bottomrule
\multicolumn{4}{l}{* `Pre-trained' means the Lepard model in SuPerPM is pre-trained }\\
\multicolumn{4}{l}{in \cite{li2022lepard}. `Fine-tuned' means the Lepard model in SuPerPM is }\\
\multicolumn{4}{l}{fine-tuned with data generated by the proposed synthesis pipeline. }\\
\multicolumn{4}{l}{* The errors are formatted as "mean (standard deviation)". The best }\\
\multicolumn{4}{l}{result in each row is in \textbf{bold}.}\\
\end{tabular}
\vspace{-1em}
\end{table}

\subsection{Implementation Details}\label{details}

To obtain the input depth maps, we employ RAFT-Stereo \cite{lipson2021raft}, a deep learning model designed to estimate the disparity map for rectified stereo images. We use RAFT-Stereo with its pre-trained weights, without further fine-tuning on our surgical datasets. 
As for tissue tracking, the procedures for initializing and adding surfels and ED nodes follow the same manner as SuPer \cite{li2020super}. 
At each new frame, we utilize the Segment Anything Model (SAM) \cite{kirillov2023segment} to segment tissue region from the background.
Finally, we set the hyperparameters for cost functions as $\lambda_{icp}=1$, $\lambda_r=10$, $\lambda_c=0.001$. 

Most hyperparameters for fine-tuning Lepard follow those from Lepard's experiments on the 4DMatch Benchmark \cite{li2022lepard}, except we downsample the point cloud ($\sim$200k) to 10k points and adjust several radius values for matching and subsampling. Full details are available with the code release.

\subsection{Results and Discussion}

We compare SuPerPM to our baseline, SuPer \cite{li2020super}, as well as advanced methods for surgical scene deformation tracking and reconstruction: DefSLAM \cite{lamarca2020defslam} and SD-DefSLAM \cite{gomez2021sd}. 
DefSLAM and SD-DefSLAM track scenes based on sparse feature matching and may not directly track the labeled points. To obtain the motion of a specific labeled point, we average the estimated motions of its 3 nearest neighbors. 

\begin{figure*}[!t]
\vspace{0.5em}
\centering
\includegraphics[width=0.99\linewidth]{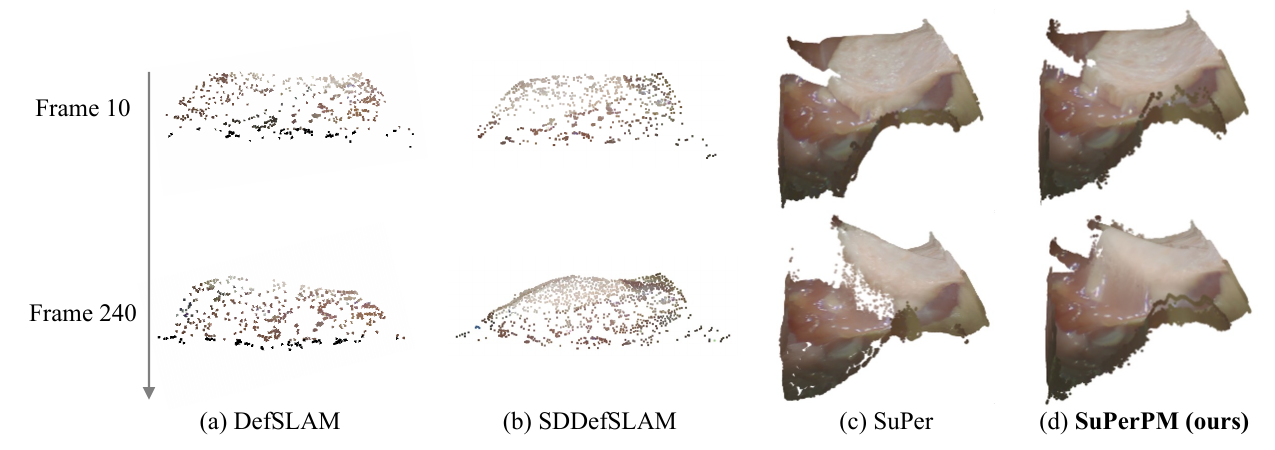}
\vspace{-1.2em}
\caption{Comparison of tracking results with SOTA methods.}
\label{fig:sample_rst}
\end{figure*}

\begin{figure}[t]
\vspace{-0.9em}
\centering
\includegraphics[width=0.5\textwidth]{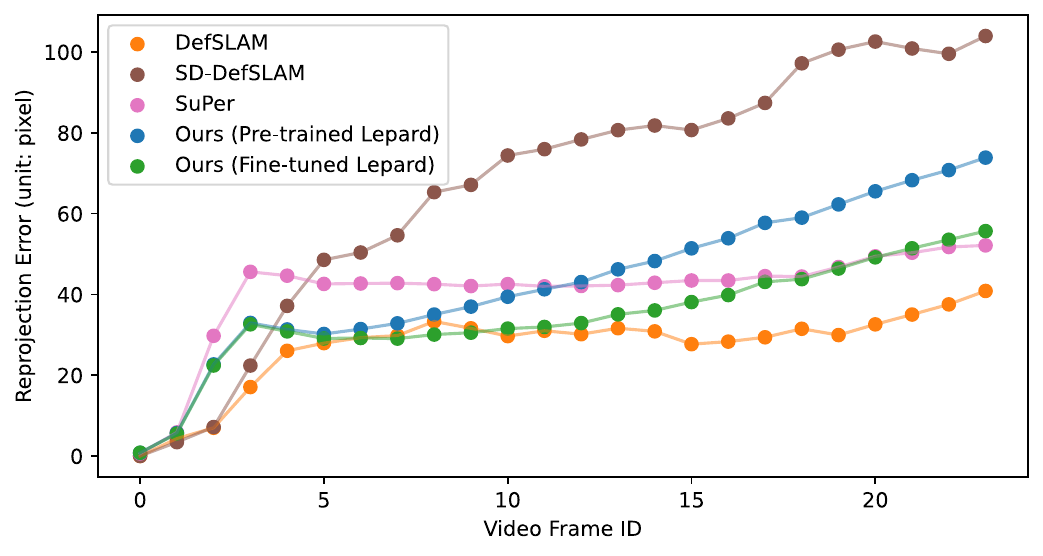}
\vspace{-2em}
\caption{Average reprojection error over time on SupDef-T2. The fine-tuned Lepard model consistently yields smaller reprojection errors compared to its pre-trained counterpart. The DefSLAM achieves the lowest reprojection errors due to only the use of sparse robust features, which is not ideal for scene reconstruction. }
\label{fig:quan_plot}
\vspace{-1.5em}
\end{figure}

We report the reprojection errors in Table \ref{tab:compare_sota}. SuPerPM surpasses its baseline, SuPer, upon which it is built. However, when replacing the ICP cost $\mathcal{L}_{icp}$ with the correspondence cost $\mathcal{L}_{corr}$, derived using the pre-trained Lepard model, there's a potential for performance degradation due to the large gap between the surgical data and data used to train Lepard, especially for data with larger deformations (SupDef-T2). 
By fine-tuning the Lepard model using data generated through the PBD-based synthesis pipeline, we significantly reduce the reprojection errors. Moreover, SuPerPM consistently outperforms SD-DefSLAM in all videos, while either matching or exceeding DefSLAM's performance. It's worth noting that both DefSLAM and SD-DefSLAM base their tracking and reconstruction on robust image features, which might result in lower reprojection errors.
However, these features are usually sparse, leading to sparse reconstruction as shown in Figure \ref{fig:sample_rst}. 
And in contrast to our results, they have difficulty accurately capturing the tissue deformations caused by grasping, which are crucial for enabling autonomous tasks by the robot. 

Figure \ref{fig:quan_plot} shows the average reprojection error at each time step in SupDef-T2, where large deformations occur throughout the entire sequence. 
In line with our findings from Table \ref{tab:compare_sota}, the fine-tuned Lepard model consistently yields smaller reprojection errors at every time step compared to its pre-trained counterpart. 
While SD-DefSLAM's performance is the worst, DefSLAM \cite{lamarca2020defslam} achieves the lowest reprojection errors on this trail, attributing to only use sparse robust image features as mentioned above.

\begin{figure}[t]
\vspace{-1.5em}
\centering
\includegraphics[width=0.5\textwidth]{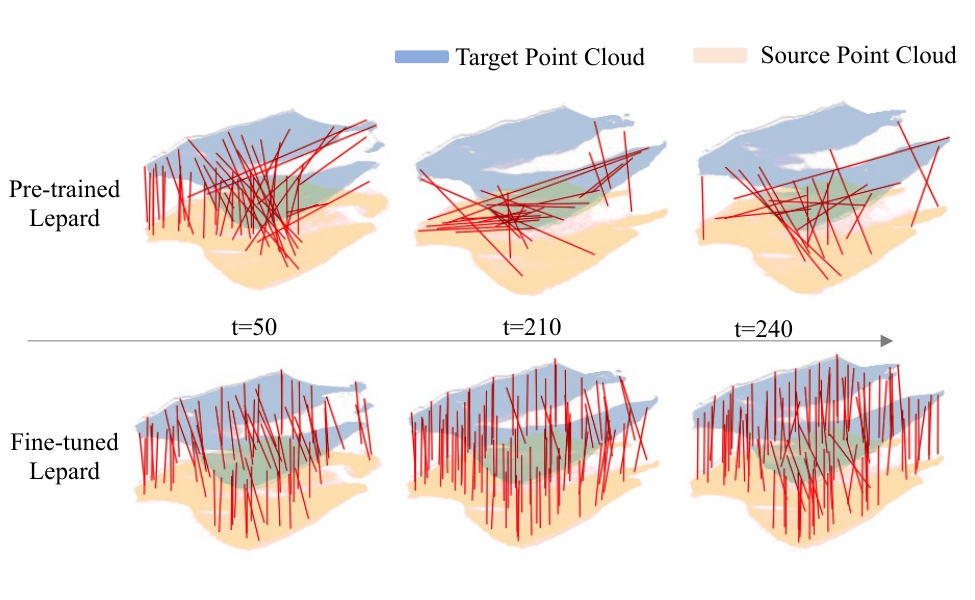}
\vspace{-3em}
\caption{Comparison of data association quality of SuPerPM across time. The pre-trained Lepard's point matching (top row) is sparse and noisy, whereas the fine-tuned Lepard (bottom row) offers denser and more consistent matches.}
\label{fig:match}
\vspace{-1em}
\end{figure}

Furthermore, we provide examples of correspondences established by both the pre-trained and fine-tuned Lepard mode during the first optimization iteration of SuPerPM at different time steps in Figure \ref{fig:match}. 
Comparing the fine-tuned Lepard model to the pre-trained Lepard model, it is evident that the fine-tuned model yields much denser and more accurate matching results. However, it is important to note that the fine-tuned model, while improved, can still produce noisy matches, as illustrated in the Figure. 
The errors can be attributed to two primary sources: 
1) While PBD is capable of generating robust correspondences by incorporating physical constraints, it may still provide incorrect correspondences due to the inherent challenges in simulating real surgical scenes; 
2) Another factor is the gap between the data used for fine-tuning and testing.  
Given that both error sources are inevitable to some extent, in the future, we plan to investigate techniques like RANSAC, geometric or physical constraints to identify and eliminate matching outliers for better tracking performance.

\section{Conclusion}
In this work, we propose a surgical perception framework SuPerPM that leverages recent advancements in deep point cloud matching.
Tissue tracking approaches rely heavily on point-wise matching to update the reconstructed tissue after deformations.
To achieve better point-wise matching than the previously used ICP, we use a learning-based matching model.
The model is fine-tuned on synthetic data generated from PBD simulations \cite{Bender_2017, Fei_2021_ICRA} of deforming tissue, hence making the model more accurate in surgical scenarios.
In our current implementation, the learning model is trained separately from the surgical perception framework. 
Considering that our framework is built in a manner that permits gradient back-propagation, in the future, we intend to enhance the training process by training the matching model together with the optimization solver, allowing the correspondence learning to be achieved in an end-to-end manner. 




\clearpage
\bibliographystyle{IEEEtran}
\bibliography{IEEEabrv,IEEEexample}

\end{document}